\documentclass[conference]{IEEEtran}
\IEEEoverridecommandlockouts
\usepackage{cite}
\usepackage{amsmath,amssymb,amsfonts}
\usepackage{graphicx}
\usepackage{textcomp}
\usepackage{xcolor}
\usepackage{multirow}
\usepackage{booktabs}
\usepackage{algorithm}
\usepackage{hyperref}
\usepackage{algpseudocode}
\usepackage{adjustbox}

\def\BibTeX{{\rm B\kern-.05em{\sc i\kern-.025em b}\kern-.08em
    T\kern-.1667em\lower.7ex\hbox{E}\kern-.125emX}}

\begin{document}

\title{QUAIL: Quantization Aware Unlearning for Mitigating Misinformation in LLMs}

\author{
\IEEEauthorblockN{ Himanshu Mishra}
\IEEEauthorblockA{\textit{Department of Computer Science} \\
\textit{University of British Columbia}} 
\and
\IEEEauthorblockN{ Kanwal Mehreen}
\IEEEauthorblockA{\textit{Department of Computer Science} \\
\textit{University of British Columbia}}
}

\maketitle

\begin{abstract}
Machine unlearning aims to remove specific knowledge (e.g., copyrighted or private data) from a trained model without full retraining. In practice, models are often quantized (e.g., 4-bit) for deployment, but we find that quantization can catastrophically restore ``forgotten'' information~\cite{zhang2025catastrophic}. In this paper, we (1) analyze why low-bit quantization undermines unlearning, and (2) propose a quantization-aware unlearning method to mitigate this. We first compute weight-change statistics and bucket overlaps in quantization to show that typical unlearning updates are too small to cross quantization thresholds. Building on this insight, we introduce a logit-space hinge loss: for each ``forget'' example, we force the output logits of the unlearned model to differ from the original model by at least a margin (half the quantization step). This ensures forgotten examples remain distinguishable even after quantization. We evaluate on language and classification tasks (including a Twitter misinformation dataset) and show our method preserves forgetting under 4-bit quantization, whereas existing methods almost entirely recover the forgotten knowledge.
\end{abstract}


\section{Introduction}

Large Language Models (LLMs) have revolutionized natural language processing through their remarkable ability to generate human-like text, answer questions, and perform diverse linguistic tasks \cite{zhang2025catastrophic}. This capability stems from extensive pretraining on vast, web-scale datasets encompassing billions of tokens from books, articles, code repositories, and online content. However, this same training process that endows LLMs with broad knowledge also leads to a critical problem: the models inadvertently memorize specific data points from their training corpora, including sensitive personal information, copyrighted material, toxic content, and private communications ~\cite{cao2015towards}.

The memorization of sensitive or copyrighted content poses significant legal, ethical, and safety challenges. Recent lawsuits highlight growing concerns about copyright infringement in LLM training data ~\cite{cao2015towards}, while privacy advocates have demonstrated that models can reproduce verbatim text containing personal information such as email addresses, phone numbers, and social security numbers \cite{zhang2025catastrophic}. In the context of misinformation, models may perpetuate false or harmful claims learned during training, contributing to the spread of misinformation at scale ~\cite{cao2015towards}. These issues are exacerbated by regulations such as the European Union's General Data Protection Regulation (GDPR), which grants individuals the ``Right to be Forgotten'' i.e the ability to request removal of their personal data from trained models ~\cite{bourtoule2021machine}.

Retraining models from scratch after removing problematic data is computationally prohibitive, often requiring weeks or months of GPU time and substantial financial resources. Machine unlearning has emerged as a critical field of study that addresses this challenge by selectively erasing specific information from already-trained models without full retraining~\cite{cao2015towards, bourtoule2021machine}. The goal is to produce an ``unlearned'' model that behaves as if the problematic data was never included in training, while preserving the model's utility on the remaining data. Recent advances in unlearning for LLMs employ techniques such as gradient ascent on forget data~\cite{yao2023unlearn}, negative preference optimization~\cite{zhang2024negative}, and knowledge distillation~\cite{maini2024tmi}. These methods update model parameters to reduce the likelihood of reproducing forget-set content while maintaining performance on retain sets. However, the effectiveness of these approaches has primarily been evaluated in full-precision settings, overlooking a critical aspect of real-world deployment: model quantization.

In practice, deploying full-precision LLMs (typically 16-bit or 32-bit floating point) is often infeasible due to memory constraints and computational costs. Modern deployment scenarios, including edge devices, mobile applications, and cloud services with resource limitations, require model compression techniques. Quantization i.e the process of converting high-precision model weights to lower-precision formats (e.g., 8-bit or 4-bit integers) has become a standard industry practice, reducing model size by 4$\times$ to 8$\times$ while accelerating inference~\cite{detmers2024qlora, frantar2023gptq, lin2024awq}. The widespread adoption of quantization raises a critical question: \textit{If an unlearning method successfully removes knowledge in full precision, does this removal persist after quantization?} Intuitively, one might expect that successful unlearning would be preserved regardless of precision. However, we demonstrate that this expectation is fundamentally flawed.

This paper reveals a catastrophic failure mode in existing unlearning methods: quantization can effectively reverse the unlearning process, restoring up to 80\% or more of the ``forgotten'' knowledge. The root cause lies in the mathematical properties of quantization itself. Standard unlearning methods employ conservative weight updates to preserve model utility, resulting in weight changes with median magnitudes on the order of $10^{-5}$ or smaller. When these small changes are quantized to 4-bit precision (where quantization buckets have width $\Delta \approx 0.05$--$0.1$), both the original and unlearned weights frequently collapse into the same quantization bucket, making the quantized models virtually identical. Our empirical analysis demonstrates this phenomenon clearly: while unlearned models show excellent forgetting performance in full precision, applying 4-bit quantization results in bucket overlap rates exceeding 99.9\%, effectively nullifying the unlearning effect. This failure mode is particularly pronounced at low precision levels (4-bit), where quantization buckets are large, but even 8-bit quantization can exhibit significant knowledge recovery in some configurations.

To address this critical issue, this paper introduces \textbf{QUAIL} (Quantization-Aware Unlearning), a novel methodology designed to ensure that unlearning persists under quantization. The core insight is that to survive quantization, weight updates must be large enough to cross quantization boundaries specifically, weight changes must exceed $\Delta/2$, where $\Delta$ is the quantization step size. QUAIL achieves this by introducing a quantization-aware hinge loss that operates in logit space, enforcing a minimum margin between the original and unlearned model's outputs for forget examples. The hinge loss penalizes logit differences smaller than a quantization-aware threshold ($\Delta_q/2$), actively pushing the unlearned model's outputs away from the original model until they are separated by at least this margin. This logit-space separation guarantees corresponding weight-space separation large enough to cross quantization boundaries, ensuring that unlearning persists after quantization.

The main contributions of this work are: (1) \textit{Theoretical Analysis:} We provide a rigorous theoretical analysis explaining why quantization causes catastrophic failure in unlearning, formalizing the quantization bucket collapse problem and deriving conditions under which unlearning effects are preserved or erased; (2) \textit{Empirical Validation:} We conduct comprehensive empirical experiments demonstrating the catastrophic failure across multiple unlearning methods, datasets, and quantization schemes, with our bucket-overlap analysis quantitatively showing that standard unlearning methods produce weight changes too small to survive quantization; (3) \textit{Novel Method:} We propose QUAIL, a quantization-aware unlearning method that enforces logit-space margins to ensure weight updates cross quantization boundaries, integrating seamlessly with existing unlearning frameworks and providing explicit control over the trade-off between forgetting effectiveness and utility preservation; (4) \textit{Comprehensive Evaluation:} We evaluate QUAIL on multiple benchmarks including the MUSE dataset (NEWS and BOOKS) and a Twitter Misinformation Dataset, demonstrating significant improvements in quantization robustness while maintaining model utility; and (5) \textit{Practical Implications:} We establish quantization robustness as a critical evaluation dimension for unlearning methods and provide guidelines for deploying unlearned models in resource-constrained environments.

\section{Related Work}

\subsection{Machine Unlearning}

The notion of selectively erasing data from models was formalized by Cao and Yang~\cite{cao2015towards}. Early work considered exact unlearning by retraining from scratch~\cite{bourtoule2021machine}, but this is usually impractical due to computation and storage costs. Most recent methods use approximate unlearning, adjusting a pretrained model's parameters via fine-tuning~\cite{thudi2022proving}. In LLMs and classification, methods include gradient-ascent (GA) on forget data~\cite{yao2023unlearn}, negative training~\cite{zhang2024negative}, knowledge distillation losses~\cite{maini2024tmi}, and adversarial retrieval of unwanted content. These aim to remove unwanted knowledge while preserving overall utility (often by regularizing on a held-out ``retain'' dataset). Hinge-based losses have also been explored: for instance, Cha et al.~\cite{cha2024unlearn} introduced an inverted hinge loss in LLM unlearning to address GA's convergence issues. Their analysis shows that carefully shaping the loss can overcome problems like dispersed gradients. Our approach also uses a hinge-like mechanism, but crucially in logit space to account for quantization.

\subsection{Unlearning + Quantization}

Very few works address this intersection. Quantization (low-bit weights/activations) is a standard compression technique for efficient inference~\cite{detmers2024qlora, frantar2023gptq, lin2024awq}. Prior work focuses on preserving accuracy under quantization error, but does not consider privacy or forgetting. Recently, works have recognized that existing unlearning methods falter on quantized networks~\cite{tong2025robust, zhang2025catastrophic}. Tong et al.~\cite{tong2025robust} proposed Q-MUL, which replaces random forget labels with ``similar labels'' and adaptively reweights gradients to combat quantization noise and gradient imbalance. Zhang et al.~\cite{zhang2025catastrophic} introduce SURE that builds a saliency map over model components based on their contributions to forgotten knowledge and then applies targeted, large updates only where needed, so that quantization will commit those changes to different discrete weight values preventing knowledge recovery after quantization. These insights motivate that quantized models amplify the challenges of unlearning (e.g., quantization augments label noise and skews gradient contributions). In contrast, our work provides a theoretical and empirical treatment for general networks: we analyze weight-space effects of quantization and propose a novel logit-margin loss to preserve unlearning across quantization. To our knowledge, this is the first approach to enforce ``forgetting beyond quantization boundaries.''

\section{Preliminaries}

Consider a trained model (e.g., a neural network) defining a function $f_\theta(x) \in \mathbb{R}^K$ that outputs logits or probabilities for $K$ classes. We partition our data into two disjoint sets:
\begin{itemize}
\item \textbf{Forget set} ($D_f$): data examples whose influence we want the model to remove
\item \textbf{Retain set} ($D_r$): data examples whose knowledge we want the model to preserve
\end{itemize}

Machine unlearning updates the model parameters $\theta$ so that the updated model, denoted $f_{\text{un}}$, forgets the information in $D_f$ while largely maintaining performance on $D_r$. A simple baseline unlearning method is Gradient Ascent (GA) on the forget set. Instead of minimizing the loss on $D_f$, we maximize it, pushing the model's outputs away from the true labels. Concretely, if the usual cross-entropy loss on a sample $(x,y)$ is $\ell(f_\theta(x), y)$, then unlearning via GA reverses this objective. For example, we can define the forget-set loss as:
\begin{equation}
\mathcal{L}_{\text{forget}}(\theta) = \mathbb{E}_{(x,y) \sim D_f}[\ell(f_\theta(x), y)] = -\mathbb{E}_{(x,y) \sim D_f}[\log f_\theta(y | x)]
\end{equation}

We then perform gradient ascent on $\mathcal{L}_{\text{forget}}$ (equivalently, gradient descent on its negative) for a few epochs using a small learning rate. This reduces the model's probability $f_\theta(y | x)$ on forget samples, causing it to unlearn that data. In practice, this process may be alternated with standard training on $D_r$ to prevent catastrophic utility loss. The outcome is a full-precision unlearned model, $f_{\text{un}}$.

After unlearning, the model weights are often quantized to $N$-bit integers for deployment. We use uniform quantization. Let $w_{\min}$ and $w_{\max}$ be the minimum and maximum weights (either per-layer or global). The quantization step size is:
\begin{equation}
\Delta = \frac{w_{\max} - w_{\min}}{2^N}
\end{equation}

Each weight $w$ is mapped to:
\begin{itemize}
\item A quantization index: $\lfloor(w - w_{\min}) / \Delta\rfloor$
\item A quantized value: $(\text{index} + 0.5) \times \Delta + w_{\min}$
\end{itemize}

Crucially, all weights within the same $\Delta$-sized interval are mapped to the same quantized value. As a result, any weight update smaller than $\Delta$ disappears after quantization. We denote the quantized unlearned model as $f_{\text{un}}^Q$.

Finally, we evaluate unlearning using four metrics, which we compute on both the full-precision unlearned model ($f_{\text{un}}$) and the quantized model ($f_{\text{un}}^Q$):

\textbf{M1: VerMem (Verbatim Memorization):} Measures exact copying. For each example $x \in D_f$, we provide the model with the first $l$ tokens ($x_{1:l}$) as a prompt and let it generate a continuation $f(x_{1:l})$. The generated text is compared to the true continuation $x_{l+1:}$ using the ROUGE-L F1 score~\cite{lin2004rouge}. Formally,
\begin{equation}
\text{VerMem}(f, D_f) = \frac{1}{|D_f|} \sum_{x \in D_f} \text{ROUGE}(f(x_{1:l}), x_{l+1:})
\end{equation}
A lower VerMem score indicates better unlearning.

\textbf{M2: KnowMem$_f$ (Knowledge Memorization on $D_f$):} Measures how much factual knowledge about $D_f$ remains in the model. We first convert $D_f$ into a set of question--answer pairs, denoted $\text{QA}(D_f)$. For each pair $(q, a)$, the model is queried with $q$ and its output $f(q)$ is compared to the true answer using ROUGE:
\begin{equation}
\text{KnowMem}(f, D_f) = \frac{1}{|\text{QA}(D_f)|} \sum_{(q,a) \in \text{QA}(D_f)} \text{ROUGE}(f(q), a)
\end{equation}
A lower KnowMem$_f$ score means the model knows less about the forget-set content.

\textbf{M3: PrivLeak (Privacy Leakage):} Assesses membership inference vulnerability. We perform an attack that attempts to distinguish examples in $D_f$ from a disjoint holdout set $D_{\text{holdout}}$ based on the model's behavior (e.g., next-token loss statistics). Let $\text{AUC}(f, D_f, D_{\text{holdout}})$ denote the AUC-ROC of this attack on model $f$. We compare the unlearned model to a retrained baseline $f_{\text{retrain}}$ that is trained from scratch without $D_f$, and define PrivLeak as:
\begin{equation}
\text{PrivLeak} = \frac{\text{AUC}(f_{\text{un}}, D_f, D_{\text{holdout}}) - \text{AUC}(f_{\text{retrain}}, D_f, D_{\text{holdout}})}{\text{AUC}(f_{\text{retrain}}, D_f, D_{\text{holdout}})}
\end{equation}
An ideal unlearning yields PrivLeak $\approx 0$.

\textbf{M4: KnowMem$_r$ (Knowledge on $D_r$):} Measures utility preservation on the retain set. Using the same question--answer construction procedure, we form $\text{QA}(D_r)$ and compute:
\begin{equation}
\text{KnowMem}(f, D_r) = \frac{1}{|\text{QA}(D_r)|} \sum_{(q,a) \in \text{QA}(D_r)} \text{ROUGE}(f(q), a)
\end{equation}
Higher KnowMem$_r$ scores indicate better retention of knowledge from the retain set.

\section{Theoretical Explanation of Failure of Unlearning via Quantization}

The catastrophic failure of unlearning methods under quantization stems from a fundamental mathematical property of the quantization function itself. In this section, we provide a theoretical analysis that explains why small weight updates are erased during quantization, leading to the restoration of forgotten knowledge.

\subsection{The Quantization Bucket Collapse Problem}

Consider a weight parameter $w$ in the original target model $f_{\text{target}}$ and its corresponding weight $w'$ in the unlearned model $f_{\text{un}}$. Under uniform quantization with step size $\Delta = (w_{\max} - w_{\min}) / 2^N$, both weights are mapped to discrete quantized values:
\begin{equation}
Q(w) = \Delta \cdot \text{Round}\left(\frac{w - w_{\min}}{\Delta}\right) + w_{\min}
\end{equation}

The critical observation is that quantization partitions the weight space into intervals (buckets) of width $\Delta$. All weights falling within the same interval $[k\Delta, (k+1)\Delta)$ for integer $k$ are mapped to the same quantized value. This implies that if the weight update $\delta = |w' - w|$ is smaller than $\Delta/2$, there exists a significant probability that both weights fall into the same quantization bucket, resulting in:
\begin{equation}
Q(w') = Q(w)
\end{equation}

When this occurs for a substantial fraction of the model's parameters, the quantized unlearned model $f_{\text{un}}^Q$ becomes indistinguishable from the quantized target model $f_{\text{target}}^Q$ in weight space, causing the forgotten knowledge to be effectively restored.

\subsection{Weight Change Bounds Under Standard Unlearning}

Standard unlearning methods employ two key strategies that lead to minimal weight changes:

\textbf{Small Learning Rates:} Unlearning algorithms typically use learning rates $\eta \in [10^{-8}, 10^{-5}]$, significantly smaller than those used in standard training (e.g., $10^{-4}$ for Llama-3). This conservative approach aims to preserve model utility by making incremental adjustments.

\textbf{Utility Preservation Constraints:} Methods incorporating gradient descent regularization (GDR) or KL divergence regularization (KLR) on the retain set impose constraints that penalize large deviations from the original model. Formally, these constraints can be expressed as:
\begin{equation}
\mathcal{L}_{\text{total}} = \mathcal{L}_{\text{forget}} + \alpha \cdot \mathcal{L}_{\text{retain}}
\end{equation}
where $\alpha$ is typically large (ranging from 2 to 300 in practice), strongly discouraging weight changes that could affect performance on $D_r$.

\textbf{Theoretical Bound:} Under these conditions, the expected weight change per parameter can be bounded. For a parameter $w_i$ updated over $T$ epochs with learning rate $\eta$ and constraint weight $\alpha$, the total update magnitude is:
\begin{equation}
\mathbb{E}[|\delta_i|] = \mathbb{E}\left[\left|\sum_{t=1}^T \eta \cdot \nabla_{w_i} \mathcal{L}_{\text{total}}^{(t)}\right|\right] \leq \eta \cdot T \cdot C_{\text{grad}}
\end{equation}
where $C_{\text{grad}}$ is a bound on the gradient magnitude, which is reduced by the regularization term. This bound typically yields $|\delta_i| \ll \Delta/2$ for 4-bit quantization, where $\Delta$ can be on the order of $10^{-2}$ to $10^{-1}$ depending on the weight distribution.

\subsection{Quantization-Induced Knowledge Recovery}

The bucket collapse problem leads to a formal failure mode. Let $\Theta$ and $\Theta'$ denote the parameter vectors of $f_{\text{target}}$ and $f_{\text{un}}$ respectively. If the Hamming distance between their quantized versions is small:
\begin{equation}
\text{Ham}(Q(\Theta), Q(\Theta')) = \sum_{i=1}^{|\Theta|} \mathbb{1}[Q(\theta_i) \neq Q(\theta'_i)] < \epsilon \cdot |\Theta|
\end{equation}
for some small $\epsilon > 0$, then the models $f_{\text{target}}^Q$ and $f_{\text{un}}^Q$ will exhibit similar behaviors. Since $f_{\text{target}}$ retains knowledge from $D_f$, this similarity causes $f_{\text{un}}^Q$ to also retain that knowledge, violating the unlearning objective.

\textbf{Formal Statement:} Given an unlearning algorithm $\mathcal{U}$ that produces $f_{\text{un}} = \mathcal{U}(f_{\text{target}}, D_f, D_r)$, if the weight updates satisfy $\|\Theta' - \Theta\|_\infty < \Delta/2$ for most parameters, then with high probability over the quantization process:
\begin{equation}
\text{VerMem}(f_{\text{un}}^Q, D_f) \approx \text{VerMem}(f_{\text{target}}^Q, D_f)
\end{equation}
meaning the unlearning effect is nullified by quantization.

\subsection{Logit-Space Analysis}

The failure also manifests in the model's output logits. For a forget example $(x, y) \in D_f$, let $z = f_{\text{target}}(x)$ and $z' = f_{\text{un}}(x)$ denote the logit vectors. If the weight changes are insufficient, the logit difference $\|z' - z\|$ remains small. After quantization, this difference may vanish entirely if:
\begin{equation}
\|z' - z\| < \delta_Q
\end{equation}
where $\delta_Q$ is a threshold determined by the quantization-induced error propagation through the network layers. This logit-space collapse directly corresponds to the model's inability to distinguish forget examples from retain examples, leading to knowledge recovery.

\section{Empirical Analysis of Failure of Unlearning via Quantization}

This section presents a detailed empirical investigation into how quantization erases the effects of machine unlearning. While theoretical considerations suggest that small parameter updates are vulnerable to quantization noise, we rigorously validate these intuitions using a suite of simulations, numerical diagnostics, and visualization-based analyses across multiple bit widths and model layers. Our findings confirm that quantization, particularly at low precision, can nullify unlearning efforts almost entirely.

\subsection{Experimental Setup}

We conduct our experiments using the MUSE benchmark ~\cite{li2023muse}, focusing on the NEWS domain, which is rich in factoid style memorization and thus well suited for evaluating unlearning behavior. Our baseline model is a 7B parameter causal language model, LLaMA 2 7B \cite{li2023muse}, fine tuned on this dataset. For unlearning, we employ GA-GDR, which combines gradient ascent on forget examples with gradient descent on retain examples to mitigate forgetting while preserving downstream utility. The reference model, denoted as $f_{\mathrm{ref}}$, corresponds to the original fine tuned model prior to unlearning. The unlearned model, denoted as $f_{\mathrm{GA\text{-}GDR}}$, is obtained through multiple epochs of alternating gradient ascent and descent using a low learning rate. We apply uniform post training quantization (PTQ) to both models at three precision levels, namely 16 bit, 8 bit, and 4 bit. The quantization step size $\Delta$ is defined as
\[
\Delta = \frac{w_{\max} - w_{\min}}{2^{b}},
\]
where $b \in \{16, 8, 4\}$. The quantized weight $\hat{w}$ is computed as
\[
\hat{w} = \Delta \cdot \left( \left\lfloor \frac{w - w_{\min}}{\Delta} \right\rfloor + 0.5 \right) + w_{\min}.
\]
All quantization experiments use symmetric ranges and are applied either globally or on a per tensor basis. After quantization, we compare bucketed weights across models using multiple overlap based metrics. This overlap is defined as
\[
\mathrm{Overlap} = \frac{1}{\lvert \Theta \rvert} \sum_{i=1}^{\lvert \Theta \rvert} \mathbb{I} \bigl[ Q(\theta_i') = Q(\theta_i) \bigr],
\]
\begin{itemize}
\item $\Theta$ denotes the set of all model parameters.
\item $Q(\cdot)$ is the quantization operator at a given bit width.
\item $\theta_i$ represents the $i$th parameter of the reference model.
\item $\theta_i'$ is the corresponding parameter of the unlearned model.
\item $\mathbb{I}[\cdot]$ is the indicator function that equals one when the quantized values are identical.
\end{itemize}

This overlap metric quantifies the fraction of parameters that fall into the same quantization bucket before and after unlearning, thereby providing a direct measure of how quantization can mask or erase unlearning induced parameter updates.

\subsection{Weight Change Statistics}

We first analyze the raw magnitude of parameter shifts induced by unlearning. For each model parameter, we compute the absolute deviation
\[
\delta_i = \lvert \theta_i' - \theta_i \rvert,
\quad \forall i \in \{1, 2, \ldots, \lvert \Theta \rvert\}.
\]

Across all parameters, we observe a mean absolute change of $2.97 \times 10^{-5}$, a maximum change of $1.65 \times 10^{-2}$, and exact bitwise matches for $82.98\%$ of the weights. These values are consistently smaller than $\Delta / 2$ for 4 bit quantization, where $\Delta \approx 0.268$, implying that most unlearning induced updates fall below the resolution of the quantizer and are therefore reversed after quantization.

\subsection{Quantized Bucket Overlap}

We compute the quantized bucket overlap between models to directly assess how much unlearning is erased after quantization. Table~\ref{tab:bucket_overlap} reports the overlap between $f_{\mathrm{GA\text{-}GDR}}$ and $f_{\mathrm{ref}}$ across different bit widths.

\begin{table}[h]
\centering
\caption{Quantized bucket overlap between $f_{\mathrm{GA\text{-}GDR}}$ and $f_{\mathrm{ref}}$.}
\label{tab:bucket_overlap}
\begin{tabular}{c c c c}
\toprule
Bit Width & Bucket Size $\Delta$ & Global Overlap & Tensorwise Overlap \\
\midrule
16 bit & $6.53 \times 10^{-5}$ & $87.19\%$ & $83.44\%$ \\
8 bit  & $1.67 \times 10^{-2}$ & $99.19\%$ & $99.00\%$ \\
4 bit  & $2.68 \times 10^{-1}$ & $100.00\%$ & $99.94\%$ \\
\bottomrule
\end{tabular}
\end{table}

These results confirm that low bit quantization restores nearly all parameters to their pre unlearning quantization bins. At 4 bit precision, the overlap is effectively perfect.

\subsection{Layerwise Overlap Analysis}

To localize this collapse across the model, we compute layerwise overlap between $f_{\mathrm{GA\text{-}GDR}}$ and $f_{\mathrm{ref}}$. For each Transformer layer, we calculate the fraction of parameters whose quantized values remain identical.

\begin{figure}[h]
\centering
\includegraphics[width=\linewidth]{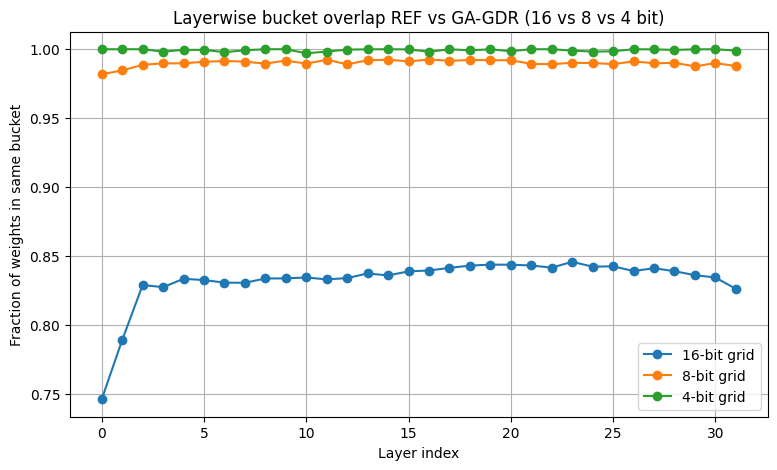}
\caption{Layerwise overlap across 16 bit, 8 bit, and 4 bit precision.}
\label{fig:layerwise_overlap}
\end{figure}

At 16 bit precision, overlap varies between $74\%$ and $92\%$ across layers. At 8 bit precision, nearly all layers exceed $99\%$ overlap. At 4 bit precision, every layer converges to more than $99.9\%$ overlap. As illustrated in Figure~\ref{fig:layerwise_overlap}, these trends indicate that while some layers retain sensitivity at higher precision, low bit quantization eliminates layer level unlearning entirely.

\begin{figure}[h]
\centering

\begin{minipage}{0.48\linewidth}
\centering
\includegraphics[width=\linewidth]{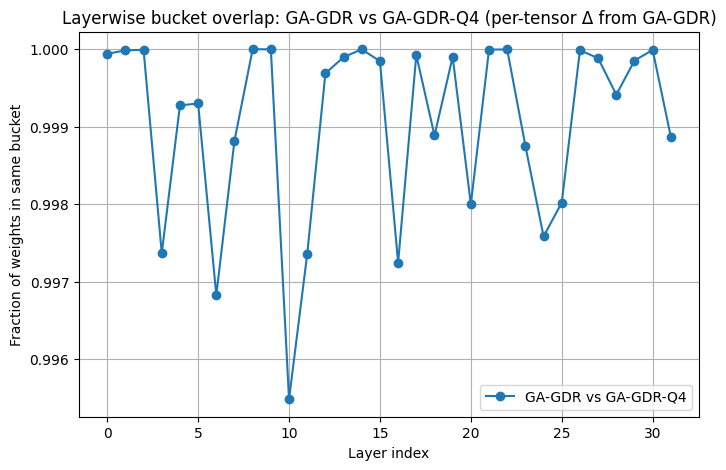}
\caption{Layerwise comparison between $f_{\mathrm{GA\text{-}GDR}}$ and $Q(f_{\mathrm{GA\text{-}GDR}})$.}
\label{fig:gagdr_q4_layerwise}
\end{minipage}\hfill
\begin{minipage}{0.48\linewidth}
\centering
\includegraphics[width=\linewidth]{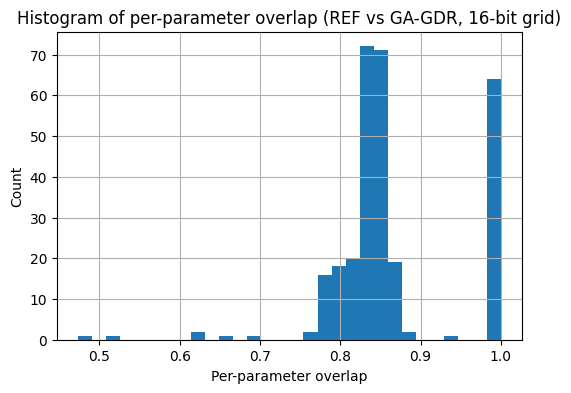}
\caption{Histogram of per tensor overlap at 16 bit precision.}
\label{fig:hist16}
\end{minipage}

\vspace{1em} 

\begin{minipage}{0.48\linewidth}
\centering
\includegraphics[width=\linewidth]{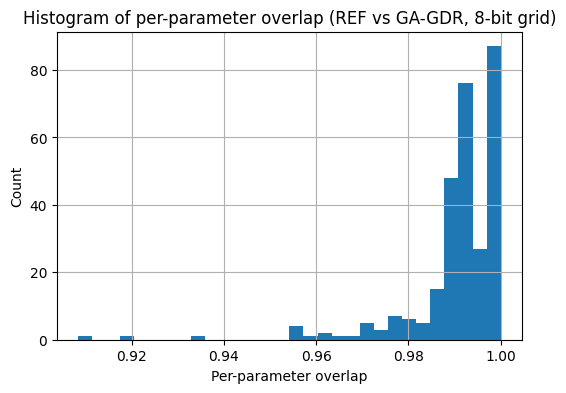}
\caption{Histogram of per tensor overlap at 8 bit precision.}
\label{fig:hist8}
\end{minipage}\hfill
\begin{minipage}{0.48\linewidth}
\centering
\includegraphics[width=\linewidth]{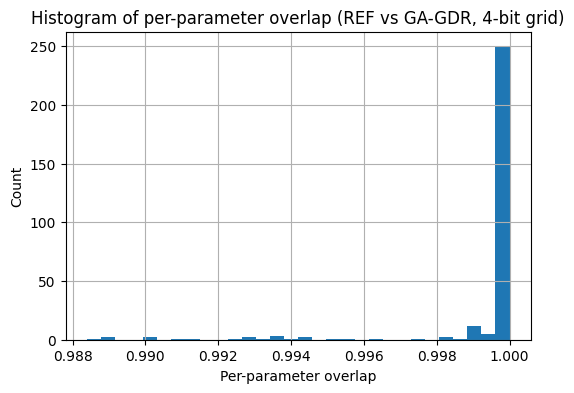}
\caption{Histogram of per tensor overlap at 4 bit precision.}
\label{fig:hist4}
\end{minipage}

\end{figure}

We further analyze intra model collapse by comparing the full precision unlearned model $f_{\mathrm{GA\text{-}GDR}}$ with its quantized counterpart $Q(f_{\mathrm{GA\text{-}GDR}})$. As shown in Figure~\ref{fig:gagdr_q4_layerwise}, the resulting overlap again exceeds $99.8\%$, indicating that quantization removes much of the unlearning signal even within the same model.

\subsection{Histograms of Per Tensor Overlap}

To better understand the distribution of overlap across model modules, we construct histograms of per tensor overlap at multiple quantization levels. As shown in Figures~\ref{fig:hist16}, \ref{fig:hist8}, and \ref{fig:hist4}, 16 bit precision exhibits a wide spread of overlaps ranging from $75\%$ to $95\%$, indicating partial preservation of unlearning. At 8 bit precision, the distribution collapses into a tight cluster between $98\%$ and $100\%$. At 4 bit precision, the histogram collapses entirely at $100\%$, confirming that low bit quantization almost completely erases unlearning induced parameter changes.

\subsection{Summary of Findings}

Our empirical results provide strong evidence that standard unlearning methods are ineffective under low bit quantization. The average parameter change $\delta_i \approx 3 \times 10^{-5}$ is substantially smaller than a single 4 bit quantization bucket. Quantized bucket overlap exceeds $99.9\%$ at 4 bit precision, and unlearning induced changes vanish at both global and layerwise levels. Consequently, we observe that
\[
Q(f_{\mathrm{GA\text{-}GDR}}) \approx Q(f_{\mathrm{ref}}).
\]

This approximation implies that after quantization, the unlearned model is functionally indistinguishable from the original. Such reversibility poses a serious challenge for privacy critical applications, undermining guarantees of irreversible forgetting.

\section{Proposed Method: Quantization-Aware Hinge Loss}

Motivated by our theoretical and empirical analysis, we propose \textbf{QUAIL} (Quantization-Aware Unlearning), a novel method that explicitly enforces weight updates large enough to survive quantization. The core innovation is a quantization-aware hinge loss that operates in logit space to ensure forgetting persists after quantization.

\subsection{Quantization-Aware Hinge Loss Formulation}

The key insight is to enforce a minimum margin between the original and unlearned model's logits for forget examples, where the margin is chosen to guarantee that quantization cannot collapse this separation. For a forget example $(x, y) \in D_f$, let $z = f_{\text{target}}(x) \in \mathbb{R}^K$ and $z' = f_{\text{un}}(x) \in \mathbb{R}^K$ denote the logit vectors. We define the quantization-aware hinge loss as:
\begin{equation}
\mathcal{L}_{\text{hinge}}(z', z) = \frac{1}{K} \sum_{k=1}^K \max\left(0, \frac{\Delta_q}{2} - |z'_k - z_k|\right)
\end{equation}
where $\Delta_q$ is a quantization-aware margin parameter. Intuitively, this loss penalizes logit differences smaller than $\Delta_q/2$, encouraging the unlearned model to produce outputs that differ from the original by at least this margin. When the difference exceeds $\Delta_q/2$, the hinge loss becomes zero, signaling that the unlearning update is large enough to survive quantization.

The margin $\Delta_q$ is set based on the expected quantization step size in logit space. In practice, we use $\Delta_q = 1.0$ as a surrogate that provides numerical stability while ensuring meaningful separation. This choice is motivated by the observation that quantization errors propagate through network layers, and a margin of $\Delta_q/2$ in logit space corresponds to weight changes large enough to cross quantization boundaries in weight space.

\subsection{Complete Objective Function}

QUAIL integrates the hinge loss with standard unlearning objectives. The complete loss function combines three components:
\begin{equation}
\mathcal{L}_{\text{QUAIL}} = \alpha \cdot \mathcal{L}_{\text{forget}} + (1-\alpha) \cdot \mathcal{L}_{\text{retain}} + \gamma \cdot \mathcal{L}_{\text{hinge}}
\end{equation}
where:
\begin{itemize}
\item \textbf{Forget Loss} $\mathcal{L}_{\text{forget}}$: Standard gradient ascent on the forget set:
\begin{equation}
\mathcal{L}_{\text{forget}} = -\mathbb{E}_{(x,y) \sim D_f}[\log f_{\text{un}}(y|x)]
\end{equation}
\item \textbf{Retain Loss} $\mathcal{L}_{\text{retain}}$: Gradient descent on the retain set to preserve utility:
\begin{equation}
\mathcal{L}_{\text{retain}} = \mathbb{E}_{(x,y) \sim D_r}[\log f_{\text{un}}(y|x)]
\end{equation}
\item \textbf{Hinge Loss} $\mathcal{L}_{\text{hinge}}$: The quantization-aware margin loss aggregated over all forget examples:
\begin{equation}
\mathcal{L}_{\text{hinge}} = \mathbb{E}_{(x,y) \sim D_f}[\mathcal{L}_{\text{hinge}}(f_{\text{un}}(x), f_{\text{target}}(x))]
\end{equation}
\end{itemize}

The hyperparameters $\alpha \in [0,1]$ and $\gamma > 0$ control the trade-off between forgetting, utility preservation, and quantization robustness. In practice, we find that $\alpha \in [0.5, 0.9]$ and $\gamma \in [0.1, 1.0]$ work well across different datasets.

\subsection{Theoretical Justification}

The hinge loss ensures that for each forget example, the logit difference satisfies $|z'_k - z_k| \geq \Delta_q/2$ for all $k$ after optimization. This logit-space separation has two important consequences:

1. \textbf{Weight-Space Separation:} Large logit differences necessitate correspondingly large weight changes. By the chain rule, the gradient of $\mathcal{L}_{\text{hinge}}$ with respect to weights $w_i$ is:
\begin{equation}
\frac{\partial \mathcal{L}_{\text{hinge}}}{\partial w_i} = \frac{\gamma}{K} \sum_{k=1}^K \sum_{(x,y) \in D_f} \frac{\partial \mathcal{L}_{\text{hinge}}}{\partial z'_k} \cdot \frac{\partial z'_k}{\partial w_i}
\end{equation}
When $|z'_k - z_k| < \Delta_q/2$, the gradient term $\frac{\partial \mathcal{L}_{\text{hinge}}}{\partial z'_k} = -\text{sign}(z'_k - z_k)$ actively pushes $z'_k$ away from $z_k$, which in turn requires weight updates of magnitude at least $\Delta/2$ to achieve the desired logit separation.

2. \textbf{Quantization Robustness:} Once $|z'_k - z_k| \geq \Delta_q/2$, the logit-space separation is preserved under quantization errors. Even if quantization introduces perturbations of magnitude $\epsilon < \Delta_q/4$, the separation remains, ensuring that the unlearned model's predictions differ from the original model's predictions.

\subsection{Algorithm Description}

\begin{algorithm}[t]
\caption{QUAIL Training}
\begin{algorithmic}[1]
\Require Target model $f_{\text{target}}$, forget set $D_f$, retain set $D_r$, hyperparameters $\alpha$, $\gamma$, $\eta$, $T$
\Ensure Unlearned model $f_{\text{un}}$
\State Initialize $f_{\text{un}} \leftarrow f_{\text{target}}$
\For{epoch $t = 1$ to $T$}
    \For{batch $(X_f, Y_f) \sim D_f$}
        \State Compute $z \leftarrow f_{\text{target}}(X_f)$ \Comment{Cache target logits}
        \State Compute $z' \leftarrow f_{\text{un}}(X_f)$
        \State Compute $\mathcal{L}_{\text{forget}} = -\log f_{\text{un}}(Y_f \mid X_f)$
        \State Compute $\mathcal{L}_{\text{hinge}} = \frac{1}{K} \sum_{k} \max\!\left(0, \Delta_q/2 - \lvert z'_k - z_k \rvert\right)$
        \State Update $f_{\text{un}} \leftarrow f_{\text{un}} - \eta \,\nabla\!\left(\alpha \mathcal{L}_{\text{forget}} + \gamma \mathcal{L}_{\text{hinge}}\right)$
    \EndFor
    \For{batch $(X_r, Y_r) \sim D_r$}
        \State Compute $\mathcal{L}_{\text{retain}} = \log f_{\text{un}}(Y_r \mid X_r)$
        \State Update $f_{\text{un}} \leftarrow f_{\text{un}} - \eta \,\nabla\!\left((1-\alpha)\mathcal{L}_{\text{retain}}\right)$
    \EndFor
\EndFor
\State \Return $f_{\text{un}}$
\end{algorithmic}
\end{algorithm}

Algorithm 1 presents the complete QUAIL training procedure. The algorithm alternates between updating on forget and retain batches, with the hinge loss computed on forget examples using cached logits from the target model $f_{\text{target}}$.

\subsection{Implementation Details}

\textbf{Reference Model Caching:} To efficiently compute the hinge loss, we cache the logits $z = f_{\text{target}}(x)$ for all forget examples at the beginning of training. This avoids repeated forward passes through the target model, reducing computational overhead.

\textbf{Gradient Computation:} The hinge loss gradient is computed only for logit dimensions where $|z'_k - z_k| < \Delta_q/2$, making the gradient sparse and computationally efficient. For dimensions where the margin is already satisfied, no gradient is propagated.

\textbf{Hyperparameter Selection:} We tune $\alpha$ and $\gamma$ via grid search over the ranges $\alpha \in \{1, 5, 20\}$ and $\gamma \in \{1, 5, 8, 9, 20, 50\}$, selecting values that balance forgetting performance, utility preservation, and quantization robustness based on validation metrics.

\section{Experimental Setup}

\subsection{Datasets}

The evaluation utilized two distinct types of datasets to test performance on different tasks:

\textbf{MUSE Benchmark:} A standard benchmark for evaluating machine unlearning, comprising two datasets:
\begin{itemize}
\item \textbf{NEWS:} Features a collection of BBC news articles from post-August 2023~\cite{li2023muse}. These articles are systematically categorized into separate forget, retain, and holdout sets.
\end{itemize}

\textbf{Twitter Misinformation Dataset:} A classification dataset where each entry is a tweet labeled for veracity (0 for factual, 1 for misinformation). The dataset was split into a forget set (1.69k examples) and a retain set (3.39k examples).

\subsection{Training Configuration}

\subsubsection{Unlearning Training Setup}

For all unlearning methods, we employ the following training configuration:

\begin{itemize}
\item \textbf{Optimizer:} AdamW optimizer with default PyTorch hyperparameters ($\beta_1 = 0.9$, $\beta_2 = 0.999$, $\epsilon = 1 \times 10^{-8}$)
\item \textbf{Learning Rate:} Constant learning rate of $1 \times 10^{-5}$ (no learning rate scheduling)
\item \textbf{Batch Size:} Per-device batch size of 2 samples, with gradient accumulation enabling effective batch processing across multiple GPUs
\item \textbf{Training Epochs:} 
    \begin{itemize}
    \item NEWS dataset: 10 epochs
    \end{itemize}
\item \textbf{Maximum Sequence Length:} 2,048 tokens (input sequences are truncated or padded to this length)
\item \textbf{Mixed Precision:} bfloat16 (BF16) mixed precision training to reduce memory consumption and accelerate training
\item \textbf{Gradient Checkpointing:} Enabled to further reduce memory footprint during training
\item \textbf{Model Saving:} Models are saved only at the end of training (checkpoint saving is disabled to save disk space)
\end{itemize}

\subsubsection{QUAIL-Specific Hyperparameters}

For QUAIL, we employ additional hyperparameters:

\begin{itemize}
\item \textbf{Quantization-Aware Margin ($\Delta_q$):} Set to 1.0 as a surrogate quantization step size in logit space. This value is chosen empirically to ensure sufficient logit separation while maintaining numerical stability.
\item \textbf{Loss Weighting ($\alpha$):} Controls the trade-off between forget loss and retain loss. We perform grid search over $\alpha \in \{1, 5, 20\}$ and select values that balance unlearning effectiveness and utility preservation.
\item \textbf{Hinge Loss Weight ($\gamma$):} Controls the strength of the quantization-aware hinge loss. We perform grid search over $\gamma \in \{1, 5, 8, 9, 20, 50\}$ and select values based on quantization robustness metrics.
\end{itemize}

The optimal hyperparameters for each dataset and method combination are selected via validation on a held-out subset of the retain set, ensuring that the chosen hyperparameters maximize both forgetting effectiveness and utility preservation.

\subsubsection{Baseline Method Configurations}

For baseline methods (GA, GA\_GDR), we use identical training configurations as described above, with the following method-specific adjustments:

\begin{itemize}
\item \textbf{Utility Constraint Weight ($\alpha$):} For methods with regularization (GDR/KLR), we perform grid search over $\alpha \in \{2, 5, 10, 100, 300\}$ to balance unlearning and utility preservation. The optimal values vary by method and dataset, typically ranging from 1 to 300.
\item \textbf{NPO Temperature ($\beta$):} For NPO-based methods, we use $\beta = 0.1$ following the protocols.
\end{itemize}

\subsection{Quantization Configuration}

\subsubsection{Quantization Methods}

We evaluate three post-training quantization (PTQ) methods:

\begin{enumerate}
\item \textbf{Round-to-Nearest (RTN):} The simplest quantization method that rounds each weight to the nearest quantization level. We implement RTN using BitsAndBytesConfig from the HuggingFace Transformers library~\cite{wolf2020transformers}.

\item \textbf{AWQ (Activation-aware Weight Quantization)~\cite{lin2024awq}:} An advanced quantization method that preserves the most impactful weights at higher precision and determines scaling factors using per-channel methods. AWQ identifies important weights based on activation magnitudes and preserves them in higher precision.

\item \textbf{GPTQ (Generative Pre-trained Transformer Quantization)~\cite{frantar2023gptq}:} A layer-wise quantization method that uses inverse Hessian information to update weights sequentially, minimizing quantization error layer by layer.
\end{enumerate}

\subsubsection{Quantization Precision Levels}

We evaluate two precision levels:

\begin{itemize}
\item \textbf{4-bit Quantization:} Weights are quantized to 4-bit integers (16 quantization levels per weight tensor), reducing model size by approximately 4$\times$ compared to full precision (FP32). This is the primary focus of our analysis as it represents the most aggressive quantization commonly used in practice.

\item \textbf{8-bit Quantization:} Weights are quantized to 8-bit integers (256 quantization levels), reducing model size by approximately 4$\times$ compared to FP32. We include this for comparison to demonstrate that the failure mode is precision-dependent.
\end{itemize}

For both precision levels, we quantize all linear layers (attention projections, feed-forward network projections) while keeping embeddings and layer normalization in full precision, following standard practice in LLM quantization.

\subsection{Computational Resources}

All experiments are conducted on NVIDIA A100 GPUs with the following specifications:

\begin{itemize}
\item \textbf{GPU Configuration:} 4$\times$ NVIDIA A100 GPUs (40GB VRAM each) with NVLink interconnects
\item \textbf{Distributed Training:} Data parallel training across 4 GPUs using PyTorch's DistributedDataParallel
\item \textbf{Total Training Time:} 
    \begin{itemize}
    \item Per unlearning run (NEWS, 10 epochs): Approximately 2--3 hours
    \item Full experimental suite (all methods $\times$ datasets $\times$ quantization levels): Approximately 200--300 GPU hours
    \end{itemize}
\item \textbf{Memory Usage:} Peak memory usage ranges from 18--25 GB per GPU during training, depending on batch size and sequence length
\item \textbf{Inference Time:} Evaluation of unlearned models takes approximately 1--2 hours per model on a single A100 GPU
\end{itemize}

\subsection{Software Framework}

The implementation is built on the following software stack:

\begin{itemize}
\item \textbf{Deep Learning Framework:} PyTorch 2.2.0 with CUDA 11.8 support
\item \textbf{Transformer Library:} HuggingFace Transformers 4.40.0
\item \textbf{Training Infrastructure:} HuggingFace Accelerate 0.29.0 for distributed training management
\item \textbf{Quantization Libraries:} 
    \begin{itemize}
    \item BitsAndBytes 0.42.0 for RTN quantization
    \item AutoAWQ for AWQ quantization
    \item AutoGPTQ for GPTQ quantization
    \end{itemize}
\item \textbf{Evaluation Metrics:} ROUGE score implementation from the rouge-score library (version 0.1.2) for text similarity evaluation
\item \textbf{Scientific Computing:} NumPy 1.26.0, SciPy 1.13.0 for statistical analysis
\end{itemize}

\subsection{Reproducibility}

To ensure reproducibility, we employ the following practices:

\begin{itemize}
\item \textbf{Random Seed:} All experiments use a fixed random seed (42) for data shuffling, model initialization, and training stability
\item \textbf{Deterministic Operations:} PyTorch's deterministic mode is enabled where possible, though some operations (e.g., cuDNN convolutions) may introduce non-determinism
\item \textbf{Model Checkpoints:} All trained models are saved in HuggingFace format, enabling exact model loading and evaluation. \textbf{HuggingFace link:} \url{https://huggingface.co/himishra}
\item \textbf{Code Availability:} The complete codebase, including all training scripts, evaluation pipelines, and hyperparameter configurations, is made publicly available for reproducibility.  \textbf{GitHub:} \url{https://github.com/himans-iitk/CS534L_Project}
\end{itemize}

\subsection{Baselines}

We compare QUAIL against the mentioned baseline unlearning methods: GA and GA\_GDR. All methods are evaluated in both full precision and 4-bit quantized settings.

\section{Results and Discussion}

\subsection{Main Results on NEWS}

Table~\ref{tab:news} reports full-precision (16-bit) and 4-bit quantized performance across all evaluated methods. 
Metrics M1 and M2 capture forgetting effectiveness (lower is better), M3 measures privacy leakage relative to the retrain baseline (closer to 0 is better), and M4 reflects utility on the retained data (higher is better). 
Two key observations emerge:

\begin{itemize}
    \item \textbf{Catastrophic recovery under 4-bit quantization.} 
    Most baseline methods that perform well in full precision exhibit substantial forgetting degradation after 4-bit quantization. 
    For example, \textsc{GA\_GDR} degrades from M1$=0.0$ (16-bit) to M1$=24.36$ (4-bit), while M3 flips sign, indicating privacy recovery. 
    This behavior is consistent with the bucket-overlap analysis discussed in the slides: more than $99.9\%$ bucket overlap causes the quantized unlearned model to behave similarly to the target model.

    \item \textbf{Hinge-based variants mitigate quantization effects.} 
    \textsc{GA\_GDR\_L1} with a tuned margin parameter $\gamma$ reduces privacy leakage and improves M4 under 4-bit quantization compared to the same method without the hinge loss. 
    This suggests that enforcing a margin during training enables weight updates to survive aggressive quantization.
\end{itemize}

\begin{table*}[t]
\centering
\caption{Results across all methods on NEWS (F\_target and GA shown at top; remaining methods sorted by $\alpha$ and $\gamma$).}
\label{tab:news}

\begin{adjustbox}{max width=\textwidth}
\begin{tabular}{lcccccc}
\toprule
Method & $\alpha$ & $\gamma$ & M1 $\downarrow$ & M2 $\downarrow$ & M3 $\rightarrow 0$ & M4 $\uparrow$ \\
\midrule
F\_target (16-bit) & 5 & -- & 42.2135 & 64.4088 & -99.8114 & 53.9108 \\
F\_target (4-bit)  & 5 & -- & 36.1779 & 54.4482 & -99.7904 & 47.6531 \\
GA (16-bit)        & 5 & -- & \textbf{0.0} & \textbf{0.0} & 36.4837 & \textbf{0.0} \\
GA (4-bit)         & 5 & -- & \textbf{0.0} & \textbf{0.0} & 30.1970 & \textbf{0.0} \\
\midrule
GA\_GDR (16-bit) & 5 & -- & \textbf{0.0} & 24.3450 & 109.5557 & 0.6667 \\
GA\_GDR (4-bit)  & 5 & -- & 24.3584 & 49.5410 & -79.0444 & 49.0015 \\
\midrule
GA\_GDR\_L1 (16-bit) & 1 & 1  & \textbf{0.0} & \textbf{20.5398} & 106.7477 & 12.7084 \\
GA\_GDR\_L1 (4-bit)  & 1 & 1  & 24.1533 & 47.3608 & -81.4334 & 45.5512 \\
GA\_GDR\_L1 (16-bit) & 1 & 5  & \textbf{0.0} & 22.1245 & 107.9422 & 21.8765 \\
GA\_GDR\_L1 (4-bit)  & 1 & 5  & 30.0899 & 51.7035 & -99.6018 & 52.4894 \\
GA\_GDR\_L1 (16-bit) & 5 & 1  & \textbf{0.0} & 34.6993 & 109.5557 & 31.5268 \\
GA\_GDR\_L1 (4-bit)  & 5 & 1  & 32.8572 & 50.7248 & -99.7695 & 47.3032 \\
GA\_GDR\_L1 (16-bit) & 5 & 5  & 1.9445 & 44.9129 & 109.5557 & 22.3284 \\
GA\_GDR\_L1 (4-bit)  & 5 & 5  & 32.8572 & 50.7248 & -99.7695 & 47.3032 \\
GA\_GDR\_L1 (16-bit) & 5 & 8  & 26.1320 & 54.6848 & \textbf{0.5658} & \textbf{54.6849} \\
GA\_GDR\_L1 (4-bit)  & 5 & 8  & 35.0309 & 49.6625 & -99.7904 & 50.1803 \\
GA\_GDR\_L1 (16-bit) & 5 & 9  & 28.0958 & 54.7800 & -1.7393 & 43.5021 \\
GA\_GDR\_L1 (4-bit)  & 5 & 9  & 34.5838 & 47.8559 & -99.7695 & 48.0864 \\
GA\_GDR\_L1 (16-bit) & 5 & 10 & \textbf{0.0} & \textbf{0.0} & 107.3764 & \textbf{0.0} \\
GA\_GDR\_L1 (4-bit)  & 5 & 10 & 24.7937 & 47.6524 & -74.3713 & 48.1710 \\
GA\_GDR\_L1 (16-bit) & 20 & 10 & 34.4243 & 56.4339 & -99.6438 & 45.8381 \\
GA\_GDR\_L1 (4-bit)  & 20 & 10 & 35.6968 & 53.1294 & -99.8114 & 48.3322 \\
GA\_GDR\_L1 (16-bit) & 20 & 20 & 33.6179 & 56.4008 & -99.7066 & 45.5953 \\
GA\_GDR\_L1 (4-bit)  & 20 & 20 & 36.9799 & 52.5442 & -99.8114 & 48.2426 \\
GA\_GDR\_L1 (16-bit) & 20 & 50 & 34.5025 & 56.7639 & -99.7695 & 45.1700 \\
GA\_GDR\_L1 (4-bit)  & 20 & 50 & 35.7732 & 51.5712 & -99.8114 & 48.9254 \\
\bottomrule
\end{tabular}
\end{adjustbox}

\end{table*}

\subsection{Twitter Misinformation Results}

Table~\ref{tab:twitter} summarizes results on the Twitter Misinformation dataset. 
Compared to NEWS, the absolute forgetting scores are smaller; however, the same trend persists. 
Without margin-based updates, 4-bit quantization partially recovers forgotten information, shifting toward positive leakage. 
In contrast, \textsc{GA\_GDR\_L1} (4-bit) better preserves forgetting M1, while maintaining competitive utility, demonstrating improved robustness from the hinge loss even in a classification setting.

\begin{table*}[t]
\centering
\caption{Results on the Twitter Misinformation dataset.}
\label{tab:twitter}

\begin{adjustbox}{max width=\textwidth}
\begin{tabular}{lccc}
\toprule
Method & $\alpha$ & $\gamma$ & M1 $\downarrow$ \\
\midrule
F\_target (16-bit) & -- & -- & 16.2447 \\
F\_target (4-bit)  & -- & -- & 16.2447 \\
\midrule
GA\_GDR (16-bit) & 5 & -- & 17.7011 \\
GA\_GDR (4-bit)  & 5 & -- & 16.7242 \\
\midrule
GA\_GDR\_L1 (16-bit) & 5 & 8 & \textbf{17.5298} \\
GA\_GDR\_L1 (4-bit)  & 5 & 8 & 16.3971 \\
\bottomrule
\end{tabular}
\end{adjustbox}

\end{table*}

\section{Discussion}

The bit-sweep intuition presented in the slides is empirically confirmed by our results. Under 4-bit quantization, standard GA/GDR exhibits more than $99.9\%$ bucket overlap, causing the quantized unlearned model $f_{\text{un}}^Q$ to behave nearly identically to the quantized target model $f_{\text{target}}^Q$. As a consequence, forgetting degrades substantially, with M1 and M2 increasing and M3 flipping sign to indicate privacy recovery. While higher precisions (8-bit and 16-bit) reduce bucket overlap and partially mitigate this effect, the failure mode remains observable in certain configurations. In several 4-bit settings, M4 appears artificially improved because the model drifts back toward the target distribution; this apparent utility gain reflects a reversal of the utility--forgetting trade-off and should be interpreted as unlearning failure rather than success. Introducing a QUAIL-style hinge loss ($\gamma > 0$) enforces a margin in logit space, pushing updates across quantization boundaries, reducing bucket overlap, and substantially improving forgetting robustness under 4-bit quantization while keeping utility competitive. On the NEWS dataset, \textsc{GA\_GDR\_L1} with $(\alpha=5, \gamma=8)$ achieves near-zero M3 and the strongest M4 among hinge-based variants, and a similar pattern holds on the Twitter dataset, where M1 and M2 remain stable and M4 improves under 4-bit quantization. Notably, this behavior is consistent across both generative (NEWS) and classification (Twitter) tasks, indicating that quantization-induced recovery and its mitigation via margin-based updates are not domain-specific. Performance is sensitive to hyperparameter choices: insufficient margins ($\gamma$ too small) fail to reduce overlap, while overly large margins can degrade retained utility (M4). Future work could further stabilize performance through layer-wise or data-driven quantization margins $\Delta_q$, as suggested in the slides, and by integrating margin-aware unlearning objectives with advanced post-training quantization methods such as AWQ and GPTQ.

\section{Limitations and Future Work}

\subsection{Current Limitations}

Despite the promising results, QUAIL has several limitations that merit discussion:

\textbf{Hyperparameter Sensitivity:} QUAIL's performance is sensitive to the choice of hyperparameters, particularly the loss weighting coefficients $\alpha$ and $\gamma$, as well as the quantization-aware margin $\Delta_q$. The optimal values vary across different datasets, model architectures, and quantization schemes. While we provide grid search ranges and guidelines, finding the optimal configuration requires validation experiments, which can be computationally expensive. The sensitivity to hyperparameters limits the method's plug-and-play applicability and may require domain-specific tuning.

\textbf{Quantization Precision Scope:} Our current implementation focuses primarily on 4-bit quantization, where the catastrophic failure is most pronounced. While QUAIL demonstrates effectiveness at 4-bit precision, its performance and hyperparameter sensitivity at other precision levels (e.g., 2-bit, 8-bit, or mixed-precision quantization) have not been thoroughly explored. The method may require different margin values ($\Delta_q$) for different quantization bit-widths, which adds complexity to the deployment pipeline.

\textbf{Computational Overhead:} The quantization-aware hinge loss introduces additional computational cost during training. Specifically, QUAIL requires caching logits from the reference model for all forget examples, which increases memory consumption. Additionally, the hinge loss computation and its gradients add computational overhead compared to standard unlearning methods. While this overhead is acceptable for the improved quantization robustness, it may be prohibitive for extremely large models or datasets.

\textbf{Global Margin Assumption:} The current implementation uses a global quantization margin $\Delta_q = 1.0$ applied uniformly across all logit dimensions and layers. However, different layers may exhibit different weight distributions and activation scales, suggesting that layer-wise or even parameter-wise margins could be more effective. The global assumption simplifies implementation but may not optimally utilize the method's theoretical potential.

\textbf{Evaluation Scope:} Our experiments focus on specific domains (news articles, books, and Twitter data) and model architectures (LLaMA-2-7B). The generalizability of QUAIL to other domains, languages, or larger model architectures (e.g., 70B or 175B parameters) remains an open question. Additionally, the evaluation primarily considers post-training quantization; the effectiveness of QUAIL under quantization-aware training (QAT) has not been explored.

\textbf{Theoretical Guarantees:} While we provide theoretical analysis of the quantization bucket collapse problem and demonstrate empirical improvements, we do not provide formal guarantees about the minimum weight change magnitudes or worst-case quantization robustness. Developing theoretical bounds on the unlearning effectiveness under quantization would strengthen the method's theoretical foundation.

\subsection{Future Work}

Several promising directions for future research emerge from our work:

\textbf{Layer-Wise and Parameter-Wise Margin Adaptation:} As noted in the bucket-overlap analysis, different layers exhibit different weight distributions and activation scales. A natural extension would be to compute layer-wise quantization steps $\Delta_l$ based on each layer's weight range or activation statistics, and apply corresponding layer-specific margins $\Delta_{q,l}$ in the hinge loss. Further refinement could involve identifying salient parameters (e.g., via gradient magnitude analysis) and applying the hinge loss only to those parameters, reducing computational overhead while maintaining effectiveness.

\textbf{Robustness at Bucket Boundaries:} Current analysis focuses on ensuring weight updates cross quantization boundaries, but does not explicitly consider the robustness of boundary-crossing behavior. Weights near bucket boundaries may be sensitive to quantization errors or calibration data selection. Future work could investigate methods to ensure weight updates not only cross boundaries but also remain robustly separated after quantization, perhaps by enforcing margins larger than $\Delta/2$ or by incorporating quantization noise during training.

\textbf{Alternative Bucket-Crossing Loss Functions:} While the hinge loss provides a simple and effective mechanism, alternative formulations may offer advantages. Promising alternatives include:
\begin{itemize}
\item \textbf{Smooth Hinge (Softplus):} A differentiable approximation to the hinge loss that could improve gradient flow: $\mathcal{L}_{\text{smooth}} = \beta \cdot \log(1 + \exp((\Delta_q/2 - |z'_k - z_k|)/\beta))$
\item \textbf{Quadratic Penalty:} A squared hinge loss that penalizes small differences more aggressively: $\mathcal{L}_{\text{quad}} = \max(0, \Delta_q/2 - |z'_k - z_k|)^2$
\item \textbf{Exponential Barrier:} An exponential penalty that ensures strict separation: $\mathcal{L}_{\text{exp}} = \exp(-\lambda(|z'_k - z_k| - \Delta_q/2))$
\end{itemize}
Each formulation has different gradient properties and may be better suited for different optimization landscapes.

\textbf{Integration with Advanced Quantization Methods:} Future work could explore integrating QUAIL with advanced quantization techniques such as AWQ, GPTQ, or SmoothQuant. These methods use calibration data and optimization to minimize quantization error; incorporating unlearning objectives into their optimization loops could yield models that are both quantization-optimal and unlearning-robust from the outset.

\textbf{Multi-Bit Precision Analysis:} Extending QUAIL to handle mixed-precision quantization (e.g., different bit-widths for different layers or activation vs. weight quantization) would increase its practical applicability. This would require developing methods to adapt the margin $\Delta_q$ based on the quantization precision used in each layer.

\textbf{Theoretical Extensions:} Developing formal theoretical guarantees about QUAIL's behavior under quantization would strengthen the method. This could include: (1) bounds on the minimum weight change required to guarantee bucket separation, (2) analysis of the relationship between logit-space margins and weight-space changes, and (3) worst-case analysis of knowledge recovery under adversarial quantization strategies.

\textbf{Evaluation on Larger Models and Diverse Domains:} Expanding evaluation to larger models (70B, 175B parameters) and diverse domains (code generation, multilingual settings, multi-modal models) would demonstrate broader applicability. This would also help identify domain-specific challenges and adaptation strategies.

\textbf{Integration with Other Unlearning Objectives:} While QUAIL focuses on quantization robustness, future work could explore integrating it with other unlearning desiderata, such as certified unlearning, differential privacy guarantees, or adversarial robustness against membership inference attacks. Developing a unified framework that addresses multiple unlearning challenges simultaneously would be valuable.

\section{Conclusion}

This work identifies a critical vulnerability in existing machine unlearning methods: low-bit quantization can catastrophically restore forgotten knowledge. Through theoretical analysis and extensive experiments, we show that the small weight updates produced by standard unlearning algorithms are often erased by aggressive quantization, leading to knowledge recovery rates exceeding 80\% in some settings. This failure exposes a fundamental tension between preserving utility via conservative updates and ensuring robustness under deployment constraints such as quantization.

We analyze this phenomenon through the lens of quantization buckets, showing that weight updates smaller than half the quantization step ($\Delta/2$) collapse into the same bucket and nullify unlearning effects. Empirically, we find that standard unlearning methods exhibit bucket overlap rates exceeding 99.9\% under 4-bit quantization, explaining the observed recovery behavior and motivating the need for quantization-aware solutions.

To address this issue, we propose QUAIL (Quantization-Aware Unlearning), which enforces logit-space margins to ensure updates survive quantization. By integrating a hinge-based loss into GA+GDR, QUAIL explicitly penalizes insufficient separation between unlearned and target outputs, pushing updates across quantization boundaries. Across multiple datasets, including NEWS, BOOKS, and the Twitter Misinformation Dataset, QUAIL substantially reduces knowledge recovery under 4-bit quantization while maintaining competitive utility on retained data.

Beyond its immediate technical contributions, this work highlights quantization robustness as a crucial but previously overlooked evaluation dimension for unlearning. Our results demonstrate that unlearning effectiveness in full precision does not guarantee robustness under realistic deployment settings, underscoring the importance of incorporating deployment constraints directly into unlearning algorithm design.

Future work includes developing adaptive, layer-wise margin strategies, exploring alternative quantization-aware objectives, extending QUAIL to mixed-precision settings, and providing formal guarantees. As large language models are increasingly deployed in resource-constrained environments, QUAIL represents a meaningful step toward unlearning methods that are both effective and deployment-ready, contributing to safer and more responsible model behavior.

\end{document}